\documentclass[cameraready]{Interspeech}

\title{Articulatory Entrainment and Coordination Complexity in Spontaneous Autistic and Non-autistic Dialogue}

\author[affiliation={1}, orcid=0000-0002-1869-1124, correspondingauthor]{Thanushi}{Withanage}
\author[affiliation={1}]{Carol}{Espy-Wilson}
\author[affiliation={2}]{Elizabeth}{Redcay}
\author[affiliation={2}]{Desi}{Jones}
\author[affiliation={3}]{Noah}{Sasson}

\address{
    $^1$ Department of Electrical and Computer Engineering, University of Maryland College Park, USA \\
    $^2$ Department of Psychology, University of Maryland College Park, USA \\
    $^3$ Department of Psychology, The University of Texas at Dallas, USA
}

\email{thanushi@umd.edu, espy@umd.com, redcay@umd.edu, drjones@umd.edu, nsasson@utdallas.edu}


\usepackage{comment}
\usepackage{csquotes}
\usepackage{subcaption}
\usepackage{amsmath}
\usepackage{float}

\begin{document}

\maketitle

\begin{abstract}
\vspace{-0.2cm}

Articulatory entrainment, the adaptation of vocal tract coordination to facilitate interaction remains underexplored in spontaneous dialogue, particularly among autistic speakers. Many prior studies have utilized task-based, phoneme-level analyses with invasive measurement techniques. Here, we introduce a speaker-independent framework to quantify articulatory entrainment in spontaneous dyadic conversations among autistic and non-autistic adults using acoustic-to-articulatory inversion and coordination complexity metrics. We examine temporal changes in articulatory coordination across interaction. Non-autistic dyads exhibit increasing coordination complexity and stronger entrainment over time, autistic dyads show moderate effects, and mixed dyads demonstrate the least alignment. Greater articulatory entrainment correlates with higher self-reported conversational success, indicating increase in coordination complexity as a marker of effective social interaction.
\end{abstract}
\vspace{-0.3cm}
\section{Introduction}
\vspace{-0.2cm}

During spontaneous social interactions, individuals tend to align with their interlocutors across multiple modalities, including acoustics \cite{willi_discriminative_2018}, prosody \cite{ochi_entrainment_2022}, facial motion \cite{sasson_non-autistic_2023}, and language \cite{patel_verbal_2022}. Such alignment has been associated with rapport, mutual understanding, and conversational success \cite{willi_discriminative_2018,lubold_conversational_2019,wynn_speech_2018,borrie_syncing_2019,levitan_measuring_2011}. Given widely held assumptions about communication challenges between autistic and non-autistic individuals, and the potential impact on well-being, characterizing how entrainment manifests across populations is essential for identifying mechanisms that support effective interaction. Understanding conversational dynamics in autism is particularly important in light of persistent claims that autistic individuals experience difficulty interpreting or expressing emotion in interaction \cite{hong_phonetic_2023} or they show atypical articulatory timing patterns linking to reduced motor speech dynamics \cite{lau_differences_2023, hong_phonetic_2022}. Rather than framing such differences as deficits, it is critical to examine the specific communicative dimensions along which autistic speakers may diverge from commonly preferred interactional norms.

Articulatory entrainment, also referred to as phonetic convergence, vocal entrainment, speech accommodation, or speech attunement, refers to voluntary or involuntary adjustments that increase similarity between conversational partners’ speaking styles and articulatory gestures. Many acoustic and phonetic work has examined convergence using non-invasive audio recordings \cite{ levitan_measuring_2011,pardo2006phonetic, pardo_comparison_2018}. However, because speech production arises from complex vocal tract dynamics, studies that directly measure articulatory kinematics have often relied on electromagnetic articulography \cite{tiede_evidence_2013, lee_articulatory_2018, vatikiotis-bateson_articulatory_2014, mukherjee_analyzing_2018} or ultrasound \cite{lawson_single_2011}, which require specialized recording setups and may alter natural speech due to sensor placement. Prior work with non-autistic participants shows that speakers can adapt tongue \cite{lee_articulatory_2018, mukherjee_analyzing_2018}, jaw, and lip movements \cite{vatikiotis-bateson_articulatory_2014} toward those of their partners. However, many articulatory and acoustic-phonetic studies use controlled or semi-structured paradigms, such as shadowing \cite{pardo_phonetic_2017}, word repetition, picture-description tasks \cite{lubold_conversational_2019}, and map-task interactions \cite{garrod1987saying}, which may not fully capture the variability of spontaneous dialogue. Thus, less is known about how articulatory coordination emerges in naturalistic spontaneous dialogue, especially in autistic and mixed-neurotype interactions.

In natural conversation, phoneme boundary detection is less reliable because coarticulation and articulatory undershoot increase speech variability \cite{lindblom_explaining_1990}. Also communicative goals especially in task-based or human–robot interactions \cite{hong_phonetic_2023} may influence normal speaking style. Research conducted with close acquaintances \cite{patel_verbal_2022, pardo_phonetic_2012} differ fundamentally from first time get-to-know-you interactions, as familiar interlocutors typically exhibit faster, more efficient, and socially attuned communication characterized by shorter conversational gaps and greater robustness to interference, whereas newly acquainted speakers engage in more tentative, deliberate, and socially cautious exchanges. Furthermore, studies including articulatory precision measures derived from read speech may not generalize to everyday interaction. Recent approaches using deep neural networks \cite{ochi_entrainment_2022} and transformer architectures \cite{lahiri_context-aware_2022} attempt to model short-term vocal entrainment by predicting interlocutors’ adjacent turns from acoustic features, yet their interpretability remains limited, making it difficult to identify which specific articulatory dimensions drive alignment. Research including autistic participants in task oriented studies have shown that autistic children and adults have reduced articulatory precision hence reduce verbal entrainment \cite{wynn_examination_2022}. Also some studies have shown that conversational convergence is different between males and females \cite{pardo_comparison_2018}.

In this study, we introduce a novel, non-invasive, speaker-independent framework to quantify the degree and direction of articulatory entrainment in spontaneous, task-free dyadic conversations using vocal tract variables derived from an acoustic-to-articulatory speech inversion technique \cite{attia_improving_2024}. We analyze articulatory entrainment in two datasets: one comprising non-autistic dyads and another including autistic, non-autistic, and mixed-neurotype interactions. We investigate how articulatory coordination emerges and evolves over time, how it differs across neurotype pairings, and how these dynamics relate to participants' perceived conversational success. By extending entrainment analysis to vocal tract dynamics, this work demonstrates that conversational alignment offer deeper insight into communication across neurodiverse interactions and informing models designed to better support effective dialogue.

\vspace{-0.2cm}
\section{Datasets}
\vspace{-0.2cm}
\subsection{CANDOR Dataset}
\vspace{-0.2cm}
Our study uses the CANDOR dataset (Conversation: A Naturalistic Dataset of Online Recordings), developed by BetterUp Labs in collaboration with the University of Pennsylvania (2023) \cite{reece_candor_2023}. The corpus includes recordings of 1,656 naturalistic, task-free dyadic 30 minute Zoom conversations. Participants were previously unacquainted adults, and after each interaction, both completed post-session surveys assessing conversational quality, from which a perceived conversational success metric was derived \cite{withanage}. Audio was recorded on separate channels per speaker, eliminating the need for diarization. Only the sessions flagged by the dataset as without substantial noise or interruptions were retained. Participants ranged in age from 19 to 66 and represented diverse gender, educational, ethnic, and generational backgrounds. From participants who completed the Autism Spectrum Quotient social skill and communication subscales; scores $\geq 6$ on either subscale were treated as elevated autistic traits, following established thresholds \cite{booth_brief_2013}. To focus on a sufficiently powered, clinically defined sample, we excluded dyads in which either participant met this elevated-trait criterion (a small, non-diagnosed subgroup), retaining 1381 non-autistic--non-autistic (NA–NA) dyadic interactions.

\vspace{-0.3cm}
\subsection{UTD Dataset}
\vspace{-0.2cm}
We chose a subset (participants who consented for their recordings to be used in studies) of the UTD dataset \cite{sasson_non-autistic_2023, jones_intersectional_2023-1} that comprises of 62 dyadic interactions, including 18 autistic--autistic (AT-AT) dyads, 31 autistic--non-autistic (AT-NA) dyads, and 13 NA-NA dyads. AT participants met diagnostic criteria on the Autism Diagnostic Observation Schedule, Second Edition (ADOS-2), and all participants had an IQ above 80.  AT participants were recruited either from the University of Texas at Dallas or from the local community. All participants were previously unacquainted and were not informed of their interlocutor’s diagnostic status. Each dyad engaged in a five-minute, unstructured, non-task-oriented, face-to-face interaction designed as a “get-to-know-you” conversation.
\vspace{-0.3cm}
\section{Method}

\vspace{-0.2cm}
\subsection{Feature extraction}
\vspace{-0.2cm}
\subsubsection{Audio prepossessing}
\vspace{-0.2cm}
For the CANDOR dataset, audio recordings were first converted from dual-channel 44.1 kHz to mono-channel 16 kHz. A 14 dB speech enhancement technique was used to reduce background noise, followed by silence removal using the Silero voice activity detection framework. The resulting speaker-specific audio streams were then segmented into 40-second chunks. 

For the UTD dataset, separate speaker audio channels for speakers were not available. Therefore, the audio feed from the centrally positioned microphone was diarized using Pyannote to obtain individual speaker audio streams. Prior to diarization, the researcher’s speech before the start of the interaction was removed from the recordings using Adobe Premiere Pro. A 14 dB speech enhancement algorithm was subsequently applied to reduce background noise, after which each speaker’s audio (without overlaps) was segmented into 3-second chunks for further analysis.
\vspace{-0.2cm}
\subsubsection{TV eigen spectra generation}
\vspace{-0.2cm}
Each audio segment was processed using a speech inversion system \cite{attia_improving_2024} to extract six time-varying vocal tract variables (TVs): lip aperture, lip protrusion, tongue tip constriction degree and location, and tongue body constriction degree and location. These articulatory trajectories were used to construct higher-order articulatory coordination feature (ACF) matrices by computing pairwise cross-correlations among TV channels with a fixed 7 second delay \cite{seneviratne_extended_2020}. For each segment, the rank-ordered eigenvalue vector (90-dimensional eigenspectrum) of the ACF matrix was computed to obtain a compact representation of articulatory coordination structure.

\vspace{-0.1cm}
\begin{figure}[h]
  \centering
  \includegraphics[width=\linewidth]{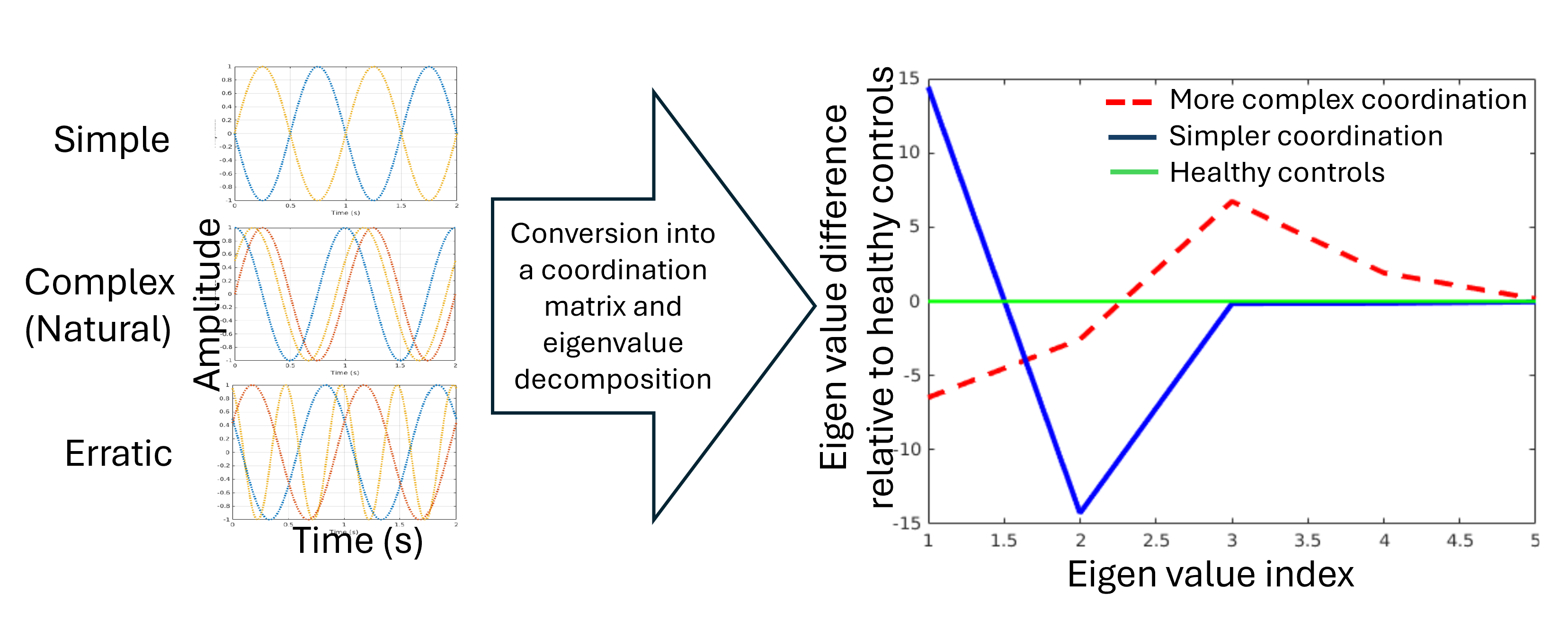}
  \caption{Coordination simulation and eigen spectra comparison for simple, complex and erratic speech}
  \label{fig:waveformcorr4}
\end{figure}
P. Divenyi and Adam Lammert \cite{divenyi2008we} simulated different types of phasing between gestures (simple, complex, and erratic) to illustrate how articulatory coordination is reflected in the eigen spectra. In the replication of those simulations shown in Figure~\ref{fig:waveformcorr4} each gesture was represented by a sine wave (first set of panels in Figure~\ref{fig:waveformcorr4}). The second plot in Figure~\ref{fig:waveformcorr4} shows that, compared to typical speech, which generally involves complex coordination, speech with simpler phasing (e.g., speech produced at a slow speaking rate) has only a few lower-ranked eigenvalues with substantial magnitude. This indicates that slower speech can be represented using a small number of independent dimensions. In contrast, more complex speech (e.g., speech produced at a faster speaking rate) exhibits more complex phasing: the lower-rank eigenvalues are smaller in magnitude relative to normal speech, while the higher-rank eigenvalues become larger. 
\begin{figure}[h]
  \centering
  \includegraphics[width=\linewidth]{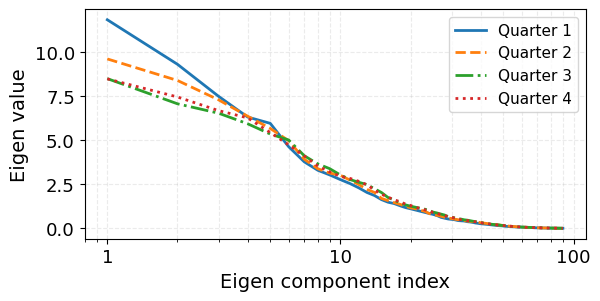}
  \caption{Eigen spectra for an individual for each quarter of the conversation}
  \label{fig:eigen2}
\end{figure}

\begin{figure}[h]
  \centering
  \includegraphics[width=0.8\linewidth,height=0.45\linewidth]{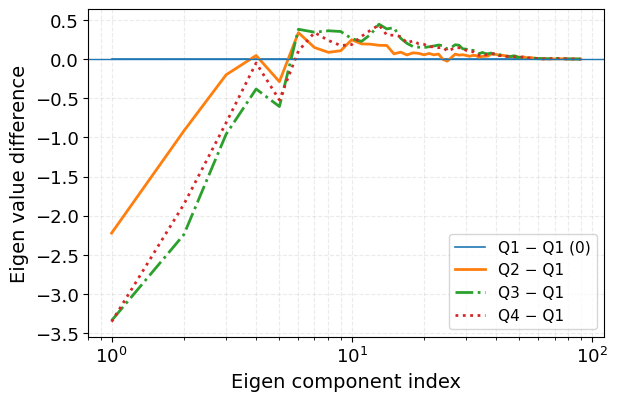}
  \caption{Difference eigen spectra for an individual for each quarter of the conversation with first quarter eigen spectra as the reference}
  \label{fig:difference3}
\end{figure}

In previous studies, we have shown that speech with simpler coordination exhibits considerably less variability (i.e., less coarticulation and undershoot) \cite{seneviratne2020classification}, whereas more complex speech shows greater variability \cite{premananth_quantifying_2025}. As pertained to this study, increased complexity may reflect greater articulatory flexibility, more relaxed speaking style and coarticulation as speakers become more comfortable, whereas reduced complexity may arise from more deliberate, effortful articulation when intelligibility is prioritized. Because the CANDOR dataset contains longer conversations, we divide each interaction into 4 quarters to examine temporal changes in articulatory coordination. As shown in Figure~\ref{fig:eigen2}, the eigen spectra of an individual progressively flatten across quarters, indicating increasing articulatory coordination complexity over time. Using the first quarter as a reference (Figure~\ref{fig:difference3}), the calculated difference eigen spectra further show that coordination complexity increases as the conversation progresses.

\vspace{-0.3cm}
\subsection{Entrainment calculation}
\vspace{-0.2cm}
In this work we define entrainment as global articulatory convergence \cite{wynn_classifying_2022} occurring in conversations. Each conversation was divided into two temporal halves and an average eigen spectrum ($v$) was calculated for each speaker taking the average of the eigen spectra of their ACF matrices within each half. This procedure was applied to both speakers in every conversation, resulting in four sets of $v$ per conversation (two speakers × two halves).

If articulatory entrainment occurs between interlocutors, the eigen spectra of the two speakers are expected to become more similar and closer as the interaction progresses. To quantify this effect, we compared the gap of the $v$ of the two speakers in the first half of the conversation and repeated the same comparison for the second half. Under the entrainment hypothesis, the gap between the speakers’ $v$ should decrease from the first half to the second half of the interaction.

A straightforward approach to measure gap is to compute the absolute distance between the two speakers’ $v$, which assigns equal weight to all eigenvalues. However, we adopted a weighted distance measure rather than a simple absolute difference to more effectively capture entrainment-related changes in eigen spectral structure. For each $v$, a weighted sum of geometric decay score ($\mathcal{W}$) was computed, similar to the approach in \cite{premananth_quantifying_2025}, as shown in Equation~\ref{equation:eq1}. In here the variable $h$ represents which half (first or second) of the conversation is used, $j$ represents which speaker (first or second) is used and $i$ represents the eigen value index in a 90 dimensional $v$. $v[i]$ corresponds to $i^{th}$ indexed term of $v$. 
\vspace{-0.3cm}
\begin{align}
\mathcal{W}_{s_j}^{(h)}
&= \sum_{i=1}^{90} v[i]_{s_j}^{(h)} \, \alpha^{i-1},
\quad j,h \in \{1,2\}, \, \alpha \in (0,1]
\label{equation:eq1}
\end{align}

This score yields a single scalar value for each $v$ by assigning greater weights to lower-ranked (more prominent) eigenvalues, with weights decaying progressively with rank. This is due to $v$ multiplied by a decay factor. The decay rate is controlled by the parameter $\alpha$, where values closer to 1 correspond to slower decay ensuring all eigen values are equally contributing whereas values closer to 0 emphasize the contribution from the most prominent eigen values. Then all $\mathcal{W} _{s_j}^{(h)}$ values were normalized to the range [0,1], yielding normalized $\mathcal{W}_{s_j}^{(h)}$ ($\omega _{s_{j}}^{(h)}$) scores. $\omega _{s_{j}}^{(h)}$ was computed for each speaker in each half of every conversation. Finally, the difference in $\omega _{s_{j}}^{(h)}$ between the two speakers ($\Delta \omega _{}^{(h)}$) was calculated for each half of the conversation as shown in Equation~\ref{equation:eq2}. A decrease in $\Delta\omega _{}^{(h)}$ from the first half to the second half was interpreted as evidence of articulatory entrainment. A Cohen’s d analysis indicated that the separation between the first and second halves' $\Delta\omega _{}^{(h)}$ is primarily driven by the most prominent eigenvalues. Specifically, varying the weighting parameter $\alpha$ showed that maximal separation in $\Delta\omega _{}^{(h)}$ occurred at $\alpha=0.51$. Accordingly, $\alpha=0.51$ was used for subsequent analyses.

To quantify entrainment, we analyze each speaker’s individual trajectory $\omega_{s_j}^{(h)}$ over time. For example, Equation~\ref{equation:eq3} computes the direction and magnitude of the shift of $s_1$ from $\omega_{s_1}^{(1)}$ to $\omega_{s_1}^{(2)}$ relative to the initial position of $s_2$, thus capturing one-way entrainment of $s_1$ towards $s_2$. Specifically, the formulation evaluates whether $\omega_{s_1}^{(2)}$ moved closer to the direction of $\omega_{s_2}^{(1)}$.

For example consider the case where the initial configuration satisfies $\omega_{s_1}^{(1)} > \omega_{s_2}^{(1)}$. Then the initial separation $\Delta \omega^{(1)}$ is negative (based on Equation~\ref{equation:eq2}), and if $s_1$ moves toward $s_2$, then the shift term $\omega_{s_1}^{(2)} - \omega_{s_1}^{(1)}$ is also negative. Consequently, both terms within the sign component of Equation~\ref{equation:eq3} are negative, yielding a positive deviation $d_1$. Thus, a positive $d_1$ indicates convergence of $s_1$(movement toward the partner $s_2$), whereas a negative value indicates divergence. The magnitude of $d_1$ reflects the strength of this directional adjustment.

Similarly, Equation~\ref{equation:eq4} quantifies the one-way entrainment of $s_2$ toward $s_1$. Directional shifts are therefore computed separately for each speaker. The overall entrainment measure $e$ is then obtained by combining the two deviations ($d_1, d_2$) and normalizing them by the initial separation $\Delta \omega^{(1)}$, as shown in Equation~\ref{equation:eq5}. This formulation yields a signed entrainment score representing total dyadic convergence or divergence as a proportion of the initial inter-speaker distance, and it remains valid across all initial configurations and directional movement scenarios.
\vspace{-0.6cm}

\begin{align}
\Delta \omega^{(h)}
&= \omega_{s_2}^{(h)} - \omega_{s_1}^{(h)},
\qquad h \in \{1,2\}
\label{equation:eq2}
\\[4pt]
d_1
&= \operatorname{sgn}(
\Delta \omega^{(1)}
(\omega_{s_1}^{(2)} - \omega_{s_1}^{(1)})
)
| \omega_{s_1}^{(2)} - \omega_{s_1}^{(1)} |
\label{equation:eq3}
\\[4pt]
d_2
&= \operatorname{sgn}(
-\Delta \omega^{(1)}
(\omega_{s_2}^{(2)} - \omega_{s_2}^{(1)})
)
| \omega_{s_2}^{(2)} - \omega_{s_2}^{(1)} |
\label{equation:eq4}
\\[6pt]
e
&= \frac{d_1 + d_2}{| \Delta \omega^{(1)} |}
= \operatorname{sgn}(\Delta \omega^{(1)})
\frac{\Delta \omega^{(1)} - \Delta \omega^{(2)}}
{| \Delta \omega^{(1)} |}
\label{equation:eq5}
\end{align}
\vspace{-0.8cm}
\section{Results}

\begin{figure*}[h]
  \centering

  \begin{minipage}{0.23\linewidth}
    \centering
    \includegraphics[width=\linewidth]{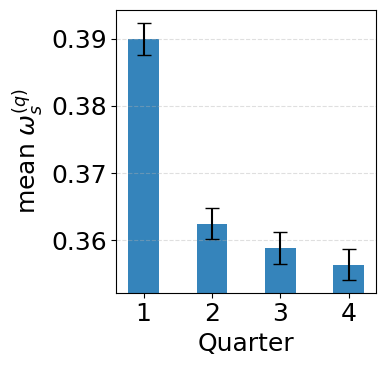}\\
    (a)
  \end{minipage}
  \hfill
  \begin{minipage}{0.25\linewidth}
    \centering
    \includegraphics[width=\linewidth]{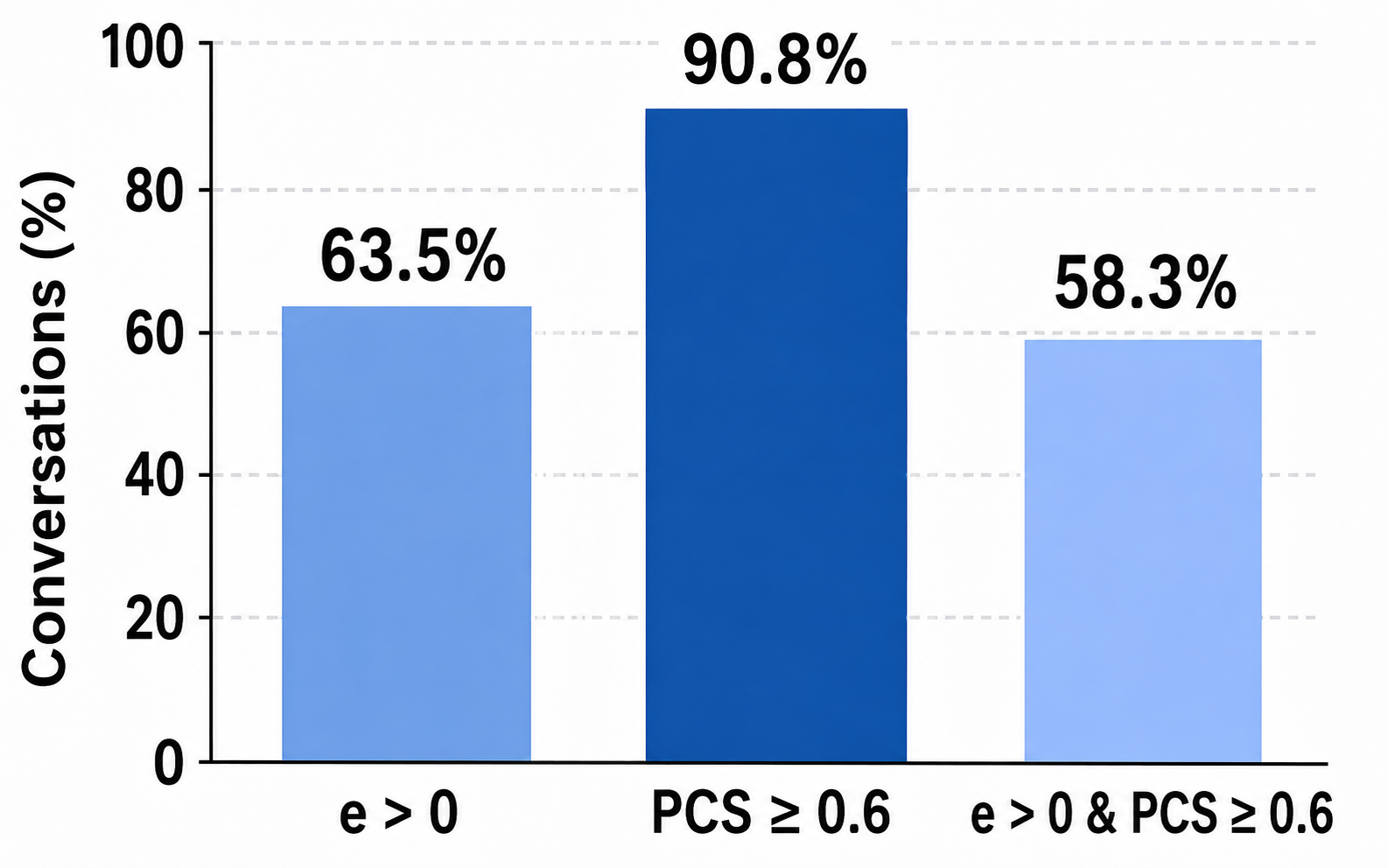}\\
    (b)
  \end{minipage}
  \hfill
  \begin{minipage}{0.27\linewidth}
    \centering
    \includegraphics[width=\linewidth]{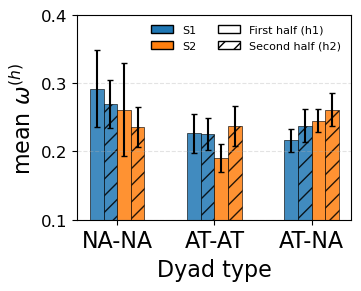}\\
    (c)
  \end{minipage}
  \hfill
  \begin{minipage}{0.22\linewidth}
    \centering
    \includegraphics[width=\linewidth]{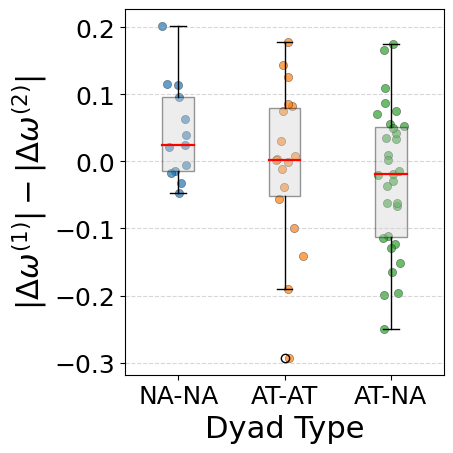}\\
    (d)
  \end{minipage}

  \caption{
  (a) Mean \& standard deviation variations of $\omega$ of NA-NA dyads vs.\ quarter of conversation for the CANDOR dataset.
  (b) Percentage of Conversations showing positive entrainment ($e>0$) and high PCS in NA-NA dyads of CANDOR dataset.
  (c) Mean \& standard deviation variations of $\omega^{(h)}$ vs.\ two halves of conversations for different dyad groups (by speaker) in the UTD dataset.
  (d) Entrainment vs.\ dyad type in UTD dataset.
  }
  \label{fig:results}
\end{figure*}

\vspace{-0.2cm}

\subsection{TV coordination complexity}
\vspace{-0.1cm}

 To observe TV coordination complexity variations over time we conducted a statistical test on the  NA-NA dyads in the CANDOR dataset utilizing $\omega _{s_{j}}^{(h)}$. A two-way repeated-measures ANOVA with half and speaker as within-subject factors showed mean $\omega _{s_{j}}$ decreased from first to second half (0.3552 to 0.3352). The analysis showed a significant main effect of half $(F(1,1380) = 170.54, p < .001)$, no significant main effect of speaker $(F(1,1380) = 0.49, p = .482)$, and no significant half $\times$ speaker interaction $(F(1,1380) = 3.07, p = .080)$; the dyad-level paired comparison showed a decrease from first to the second half $(t(1380) = 13.06, p < .001)$, while within-half speaker differences were not significant. Taken together, these results indicate a pronounced shift from the first to the second half of conversations that is shared across speakers, with negligible speaker-specific variation. Hence $\omega _{s_{j}}^{(1)}>\omega _{s_{j}}^{(2)}$ and justifies that speakers become more complex coordinated over time.

A finer-grained analysis of the mean and standard deviation of $\omega_{s_j}^{(h)}$ across conversation quarters (Figure~\ref{fig:results}(a)) reveals a marked decrease from the first to the second quarter, indicating increasing coordination complexity. To further refine this observation, a two-way repeated-measures ANOVA with quarter and speaker as within-subject factors; mean $\omega_{s_j}^{(q)}$ decreased from 0.3904 to 0.3560. The analysis showed a main effect of quarter $(F(3,4134) = 136.81, p < .001)$, no main effect of speaker $(F(1,1378) = 0.41, p = .524)$, and a quarter $\times$ speaker interaction $(F(3,4134) = 2.94, p = .032)$. FDR-corrected pairwise comparisons confirmed progressive changes across quarters, while within-quarter speaker differences were largely non-significant. As illustrated in Figure~\ref{fig:results}(a), the largest decline in mean $\omega$ for both speakers occurs early, approximately within the first 7 minutes suggesting that increased articulatory coordination emerges relatively early in the interaction.

For the UTD dataset $\omega_{}^{(h)}$ range variation of each dyad group over time are plotted in Figure ~\ref{fig:results}(c). When observing the mean $\omega$ it is clear that NA-NA dyads showed complex coordination towards the end indicating comfort,  AT-AT dyads participants maintain similar complexity throughout the conversation without much variation, whereas in mixed dyad group participants tend to get more simpler coordinated at the end. This might happen due to the thought overload of trying to make oneself more clearer and explainable toward the other interlocutor when the partner is a different neurotype. This is consistent with the double empathy framework \cite{sasson_non-autistic_2023} and is consistent with the results observed in CANDOR dataset.

\vspace{-0.3cm}
\subsection{Entrainment}
\vspace{-0.2cm}
Figure~\ref{fig:results}(b) illustrates the relationship between perceived conversational success and the metric $e$ in NA-NA dyads of CANDOR dataset. When $0<e<1$, the speakers progressively reduce their separation over time, indicating gradual entrainment toward the end of the conversation. If $e>1$, the speakers’ $\omega$ trajectories cross at some point during the interaction, implying strong convergence with overlap in the intermediate stage. An outcome of $e=0$ indicates that the speakers maintain a constant separation and stabilize at different $\omega$ values without convergence or divergence. When $e<0$, the speakers diverge over time, reflecting a lack of entrainment. The first two cases ($e>0$) are therefore interpreted as entrainment. Majority of highly successful conversations fall within the range $e>0$. This pattern suggests that articulatory entrainment is associated with higher perceived conversational success in the NA-NA dyad group.

When examining the entrainment patterns in the UTD dataset, we divided each conversation into two halves, consistent with the initial analysis conducted on the CANDOR dataset. Further segmentation was not performed because the interactions are relatively short in duration. Figure~\ref{fig:results}(d) illustrates the entrainment range, measured using $|\Delta \omega _{}^{(1)}| - |\Delta\omega _{}^{(2)}|$, for conversations grouped by dyad type. If  $|\Delta \omega _{}^{(1)}| - |\Delta \omega _{}^{(2)}| > 0$ then that conversation has entrained, while if $|\Delta \omega _{}^{(1)}| - |\Delta \omega _{}^{(2)}| < 0$ then that conversation is not entrained. Using that logic,  results shown in Figure~\ref{fig:results}(d) depicts that NA–NA dyads exhibit the highest level of entrainment (mostly positive), whereas the lowest entrainment is observed in the mixed dyad group. Interestingly AT-AT dyads have shown relatively more entrainment than mixed dyads showing that there is more alignment between same neurotype participants when conversing. This is consistent with the results observed in CANDOR dataset.

\vspace{-0.7cm}
\section{Conclusion}
\vspace{-0.2cm}
This study introduces a principled quantitative framework for measuring articulatory entrainment based on dynamic changes in coordination patterns over the course of an interaction. Articulatory features allow us to examine speech not only as an acoustic signal, but as a dynamic motor behavior that reflects how speakers coordinate, adapt in an interpretable manner. Across two independent datasets, we demonstrate that same-neurotype dyads exhibit increasing articulatory coordination becoming more complex coordinated as conversations progress, suggesting greater interactional comfort and reduced articulatory effort toward the end of the exchange. In contrast, mixed-neurotype dyads show patterns consistent with sustained or increased articulatory effort, potentially reflecting adaptive strategies to enhance speech clarity and mutual intelligibility. This is in line with the Lindblom’s Hypo- and Hyper-articulation theory. Also, articulatory entrainment is more prominent and shows a positive association with perceived conversational success among non-autistic interactions. These findings highlight the role of articulatory coordination and entrainment as markers of interactional alignment and communicative efficacy. Future work will extend this framework to datasets involving autistic children and to multi-participant interactions in workplace and task-oriented settings, enabling a broader understanding of articulatory coordination dynamics across developmental stages and social contexts.
\section{Acknowledgments}
This research was supported by the Grand Challenge grant (USource:US009754) from University of Maryland. The authors gratefully acknowledge the participants whose data were used in this study. We sincerely thank them for their time and contribution to advancing research on speech, communication, and social interaction.

\section{Generative AI Use Disclosure}
The authors used ChatGPT by OpenAI to assist only with language polishing and readability improvement. The authors reviewed and edited the manuscript after using this tool and take full responsibility for the content of the publication.
\bibliographystyle{IEEEtran}
\bibliography{refnz}

\end{document}